\documentclass[11pt]{article}

\usepackage[final]{acl}

\usepackage{eccvabbrv}

\usepackage{graphicx}
\usepackage{booktabs}

\usepackage[accsupp]{axessibility}  

\usepackage{hyperref}
\usepackage{array}
\usepackage[most,breakable]{tcolorbox}
\newcolumntype{H}{>{\setbox0=\hbox\bgroup}c<{\egroup}@{}}

\newcommand{\ours}{MobileWorld\xspace}
\newcommand{\bench}{MobileWorldBench\xspace}
\usepackage{orcidlink}
\usepackage{amssymb}
\usepackage{pifont}
\newcommand{\qwenmax}[0]{Qwen3-VL-235B-A22B}
\usepackage[table]{xcolor}
\newcommand{\cmark}{\ding{51}}%
\newcommand{\xmark}{\ding{55}}%

\usepackage{cleveref}
\usepackage[skip=3pt]{subcaption}
\usepackage{times}
\usepackage{latexsym}

\usepackage[T1]{fontenc}

\usepackage[utf8]{inputenc}

\usepackage{microtype}

\usepackage{inconsolata}

\usepackage{graphicx}

\title{MobileWorldBench: Towards Semantic World Modeling For Mobile Agents}

\author{
 \textbf{Shufan Li \textsuperscript{1}},
  \textbf{Konstantinos Kallidromitis \textsuperscript{2}},
  \textbf{Akash Gokul \textsuperscript{3}},
\textbf{Yusuke Kato \textsuperscript{2}} \\
    \textbf{Kazuki Kozuka \textsuperscript{2}},
 \textbf{Aditya Grover\textsuperscript{3}}
\\
\\
 \textsuperscript{1}UCLA  \textsuperscript{2}Panasonic  \textsuperscript{3}NVIDIA 
\\
 \small{
   \textbf{Correspondence:} \href{mailto:email@domain}{jacklishufan@cs.ucla.edu}
 }
}

\begin{document}
\maketitle 


\begin{abstract}
World models have shown great utility in improving the task performance of embodied agents. While prior work largely focuses on pixel-space world models, these approaches face practical limitations in mobile agent settings, where predicting complex visual GUI elements in future states is often difficult. In this work, we explore an alternative formulation of world modeling for mobile agents, where state transitions are described in natural language rather than predicting screenshot pixels. First, we introduce \bench, a benchmark that evaluates the ability of vision-language models (VLMs) to function as world models for mobile agents. Second, we release \ours, a large-scale dataset consisting of 1.4M samples, that improves the world modeling capabilities of VLMs. Finally, we propose a novel framework that integrates VLM world models into the planning framework of mobile agents, demonstrating that semantic world models can directly benefit mobile agents by improving task success rates. Code and data is available at \href{https://github.com/jacklishufan/MobileWorld}{https://github.com/jacklishufan/MobileWorld}.

\end{abstract}
 
\section{Introduction}
\label{sec:intro}

World models are capable of predicting future states of a system given current observations, making them particularly useful in a wide range of applications such as robotics, physical simulations,  video games, and autonomous driving \cite{mazzaglia2024multimodal,yue2025ewmbench,gu2025cosmos,guo2025mineworld,zheng2025world4drive}. The most common approach for world modeling is action-conditioned causal video modeling \cite{chen2025learning,gu2025cosmos,bruce2024genie}, where a neural network is employed to predict pixels in future video frames based on current and past frames, and action inputs. The predictions of these video world models can be naturally integrated into model-based policies, which have demonstrated strong utility both in real-world tasks and in simulated environments \cite{ren2023surfer,chen2025learning}.

\begin{figure}[t]
    \centering
    \includegraphics[width=1.0\linewidth]{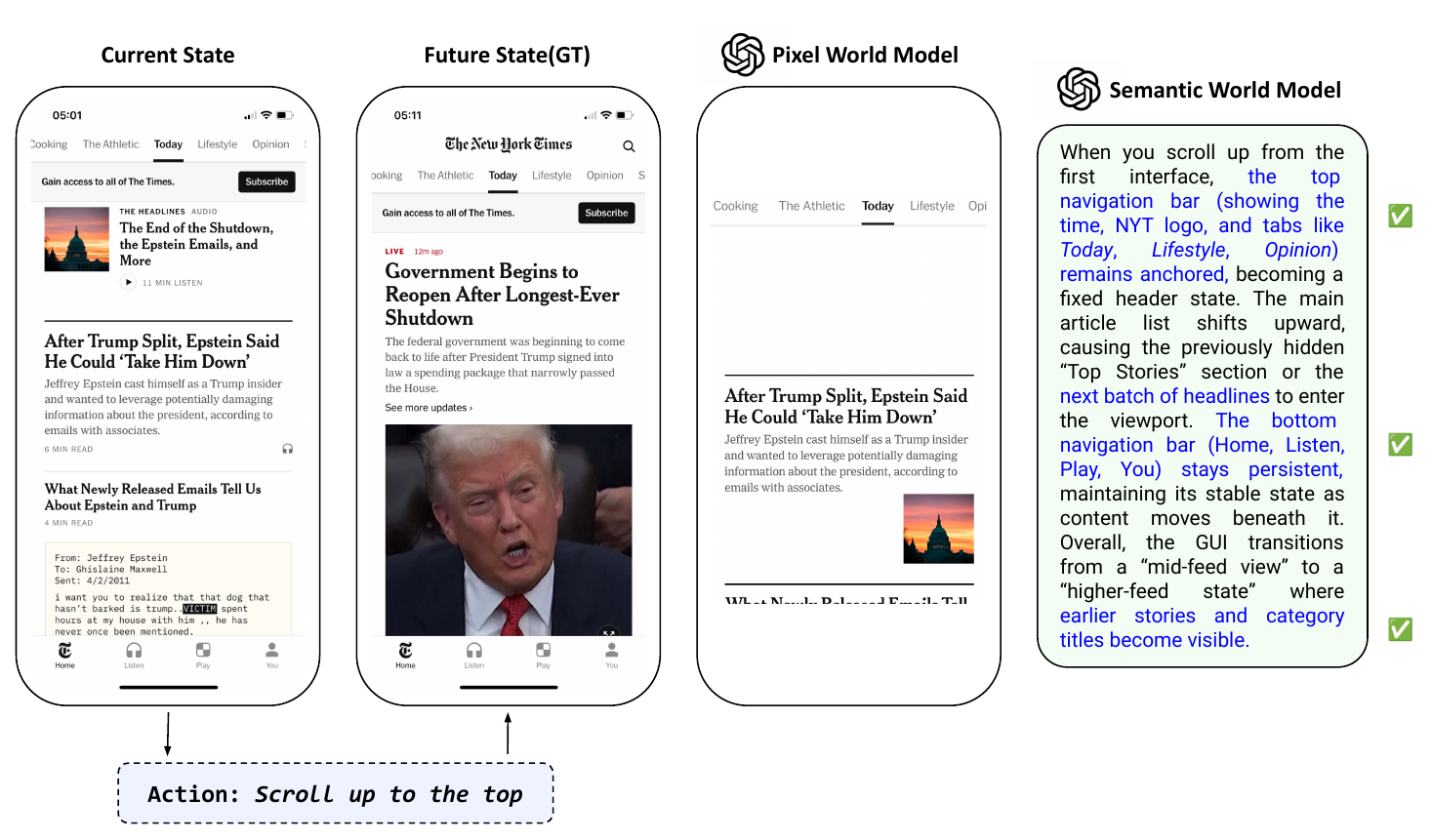}
    \caption{\textbf{Advantages of Semantic World Modeling.} Pixel-space world modeling is particularly challenging as the model needs to identify changes, come up with the correct app content, and render them accurately. By contrast, Semantic World Modeling only focuses on abstracting relevant changes in GUI semantics, while being useful for decision-making. In the example shown, the frontier visual generative model GPT-Image struggles to accurately render GUI states, while frontier VLMs (GPT-4o \cite{hurst2024gpt}) can accurately describe the expected GUI changes in text.}
    \label{fig:realteaser}
\end{figure}

Despite their success in many domains, world models for mobile agents remain under-explored. A key challenge in applying naive next-frame prediction to mobile world modeling lies in the intrinsic complexity of the pixel-level prediction task. Let's use a news app, \eg the New York Times app, as an example (Fig \ref{fig:realteaser}). Suppose a user is browsing a list of news articles, and will perform the action \textit{``scroll to the top"}. To predict the next state, a pixel-space world model needs to: 1) Understand the high-level semantics of the action, namely that the scroll action will likely result in additional articles showing up, while keeping the majority of the UI layout unchanged (\eg bottom navigation bar, top tabs, subscribe button), 2) Predict the visual layout of the future page: \eg Does the new app present thumbnails and abstracts? Or does it only show headlines?, 3) Predict the exact content, which involves generating plausible news, 4) Render all texts and icons based on the predicted content and layout.

Even if a world model masters all four capabilities, it remains highly unlikely that the predicted next state will be visually similar to the actual next state, since it is improbable that the predicted news is the same as the real news at the given time.  Similarly, it is also hard to predict if new articles include thumbnails, short summaries, or are just titles. In general, 2), 3), and 4) are exceptionally hard tasks that are challenging for state-of-the-art visual generative models. Moreover, these details are typically not essential for model-based policies. For instance, if an agent wants to perform a different action such as \textit{``subscribe"} or \textit{``navigate to view sports news"}, rendering the exact content of new articles that will show up is unnecessary. 

Rather than relying on pixel-level predictions, we hypothesize that mobile agents can achieve far more efficient and generalizable world modeling by representing state changes as structured, semantically meaningful textual representations. Notably, traces of this idea appear in the chain-of-thought behavior of large language models (LLMs), which often verbalize predictions about future states in their reasoning. Our contribution is to formalize this phenomenon as an explicit modeling framework, treating textual state descriptions as world models. In doing so, we transform what was previously a by-product of reasoning into a principled mechanism for planning and control. Concretely, this work introduces three key contributions to facilitate the paradigm of semantic world modeling for mobile agents:

First, we propose \bench, a comprehensive benchmark that explicitly evaluates VLMs' world modeling capabilities. Unlike existing GUI understanding and grounding tasks, which focus on interpreting elements on the current screen, \bench evaluates VLMs’ ability to predict future states from the current screen and a given user action. Notably, \bench involves two tasks: Next-State-Generation and Next-State-QA. In the Next-State-Generation task, the model produces free-form text describing predicted state transitions, which are then evaluated by a VLM judge that compares the description with the ground-truth screenshots of the next state. In Next-State-QA, the model answers a series of yes-no questions about future states, and its performance is quantified using an accuracy metric that directly measures world modeling ability.


Second, to facilitate the training of semantic world models for mobile agents, we curate \ours, a large-scale world modeling dataset consisting of  the triplets: \textit{current state, user action,} and \textit{future state}. Future states are represented in three forms: pixels of screenshots, QA pairs, and natural language descriptions of state transitions. We construct \ours by sourcing existing trajectory data and leveraging advanced VLMs to generate QA pairs and annotate state transitions using ground-truth next states.

Finally, we analyze the effectiveness of semantic world models by finetuning an open-source VLM using \ours. Experimental results show that mobile agents that use semantic world models perform better on the AndroidWorld \cite{rawles2023androidinthewild} benchmark ($+7.4\%$ increase in success rate).

\section{Related Works}

\subsection{World Modeling}

Prior work on world modeling has largely focused on predicting future observations either at the pixel level or in latent space. Pixel-based approaches include GUI simulators and video generation systems such as NeuralOS \cite{rivard2025neuralos}, MineWorld \cite{guo2025mineworld}, GEM \cite{hassan2025gem}, Cosmos \cite{gu2025cosmos}, Genie \cite{bruce2024genie}, and ViMo \cite{luo2025vimo}. Latent-space approaches such as V-JEPA \cite{bardes2024revisiting,assran2025v} improve efficiency by predicting compact representations instead of reconstructing pixels. More recent work explores semantic world modeling through language-based abstractions, including SWM \cite{berg2025semantic} and WMA \cite{chae2024web}, which formulate future-state prediction as textual reasoning. Other studies further demonstrate that language can serve as an effective abstraction for planning and decision-making \cite{jiang2019language,chen2025planning,chen2025worldprediction}. 

Unlike prior work that relies on pixel reconstruction, we focus directly on semantic future-state prediction. Compared with prior semantic world modeling works on web navigation, our work uniquely address the semantic world modeling for mobile GUI agents, which has more diverse tasks and requiring additional long-horizontal planning capabilities because they involves many persistent states like file systems and system setting.

\subsection{Mobile Agents}

Mobile agents have evolved from systems relying on structured UI representations \cite{wen2023droidbot,wen2024autodroid}, to multimodal agents that operate directly on screenshots \cite{zhang2025appagent}. Recent works improve decision-making using memory, decomposition, or reinforcement learning \cite{wang2024mobile,li2024appagent,ye2025mobile,bai2024digirl,liu2024autoglm}, while others enhance GUI grounding and perception through specialized architectures and supervised finetuning \cite{wang2025ui,hong2024cogagent,ma2024coco,cheng2024seeclick,lu2025guiodyssey}. Existing agents typically select actions through UI element grounding or coordinate prediction \cite{wen2023droidbot,wang2024mobile,li2024appagent} and operate in a reactive observation-action loop. In contrast, our work studies predictive modeling of future GUI states to support planning and long-horizon reasoning.

\subsection{GUI Datasets \& Benchmarks}

Existing GUI datasets and benchmarks primarily focus on interface understanding, grounding, or agent execution. Representative datasets include Rico \cite{deka2017rico}, Screen2Words \cite{wang2021screen2words}, AITW \cite{rawles2023androidinthewild}, and GUICourse \cite{chen2025guicourse}. Popular benchmarks include Mind2Web \cite{deng2023mind2web}, WebArena \cite{zhou2023webarena}, Mmina \cite{tian2025mmina}, Spotlight \cite{li2022spotlight}, Ferret-UI \cite{you2024ferret}, OmniACT \cite{kapoor2024omniact}, AgentStudio \cite{zheng2024agentstudio}, OSWorld \cite{xie2024osworld}, AndroidWorld \cite{rawles2024androidworld}, and GUI-World \cite{chen2024gui}. Unlike these works, which mainly evaluate action selection or understanding of observed states, our work focuses specifically on predicting future GUI states conditioned on user actions. 
\section{Method}
\begin{figure*}[t]
    \centering
    \includegraphics[width=1.0\linewidth]{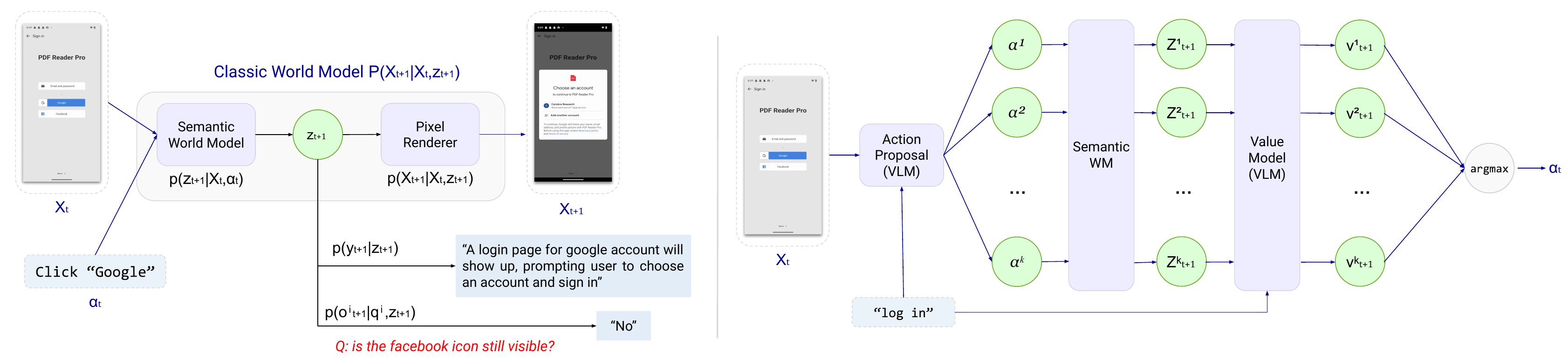}
    \caption{\textbf{The Semantic World  Model Paradigm.} (Left): we factorize the classic pixel world models into two components. We call the first component the semantic world models. It predicts the latent distribution $p(z_{t+1}|X_t,a_t)$ encoding high-level semantics. $z_{t+1}$ can be queried via $p(y|z)$ to produce text descriptions, or through $p(o^i|q^i,z)$ to produce yes-no answers. (Right). To use semantic world models for decision-making, we employ a model-based policy framework that combines a semantic WM with an action proposal model and value model.  }
    \label{fig:explainer}
\end{figure*}

\begin{figure*}[t]
    \centering
    \includegraphics[width=0.99\linewidth]{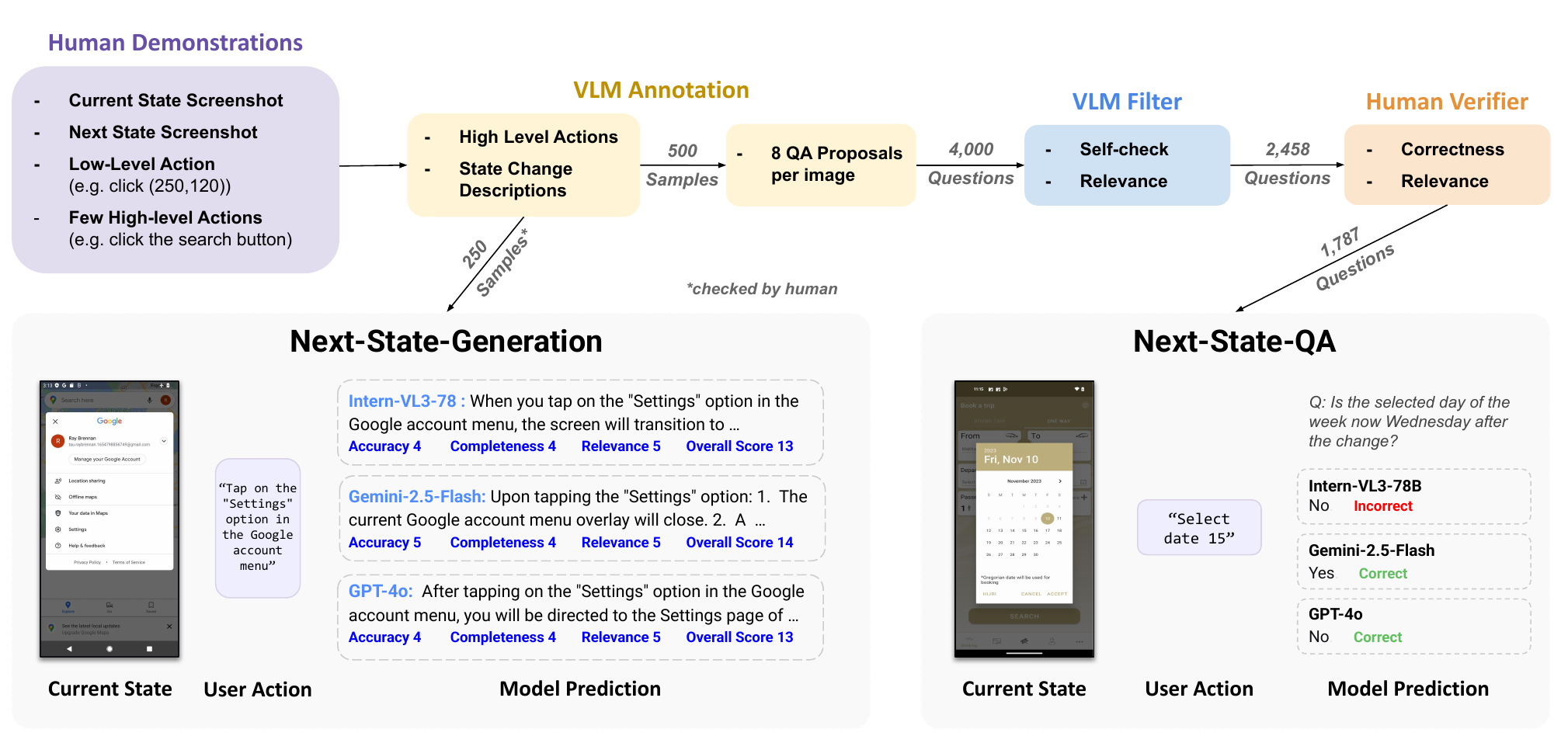}
    \caption{\textbf{Data Pipeline for \bench.} Our data pipeline consists of 5 steps: 1) Curating raw trajectories, 2) Using VLM to convert low-level actions to high-level actions and annotate state-change descriptions for reference, 3) We generate QA candidates for each state transition sampled, 4) We use VLM to filter these QA pairs through self-check and relevance metrics, 5) For the filtered data, we additionally use human verification to further filter the data based on its correctness and relevance.}
    \label{fig:placeholder}
\end{figure*}

\subsection{Semantic World Modeling for Mobile Agents}
\label{sec:formulation}
Given a current GUI state $X_t$ at timestep $t$ and a user action input $a_t$, classic world modeling $p(X_{t+1}|X_t,a_{t})$ directly predicts the pixel representation of the future state $X_{t+1}$. We break down this process into two steps. Given $(X_t,a_t)$, we first predict the high-level changes that will occur, such as ``the search bar will appear", or ``a new word is typed in the input box". We represent these changes as a latent variable $z_{t+1}$. Then we can render the pixel-level details based on the predicted changes $z_{t+1}$ and previous screen $X_t$. Formally, we factorize the classic world modeling objective $p(X_{t+1}|X_t,a_t)$ as:

\begin{align}
    p(X_{t+1}|X_t,a_t) \nonumber \\  = \sum _{z_{t+1}} p(X_{t+1}| & z_{t+1},X_t,a_{t})p(z_{t+1}|X_t,a_t)
\end{align}

where $p(X_{t+1}|z_{t+1},X_t,a_t)$ is assumed to be independent of the action $a_t$ and can be written as $p(X_{t+1}|z_{t+1},X_t)$, \ie once we know the changes that will occur, we can use it to determine the next state without knowing the action. We refer to the distribution \(p(z_{t+1}\mid X_t,a_t)\) as the semantic world model.

In practice, we use natural language descriptions to represent actions. We assume $z_{t+1}$ lies in the hidden representation of a VLM. To extract interpretable information from a latent $z$, we define two queries on the latent variable. The first query $p(y|z)$ generates a text description $y$ that summarizes the GUI changes encoded in $z$. Additionally, we also provide a series of statements $\{q^i\}_{i=1,2..N}$ about the semantics of the next state. The second query $p(o^i|q^i, z)$ can be used to obtain the likelihood of $q^i$ being true in the next GUI state. These two queries allow us to evaluate the quality of $z$ and meaningfully utilize them in decision-making. These setups are illustrated in \cref{fig:explainer} (Left).

To utilize semantic world models for decision-making, we adopt a model-based policy approach shown in \cref{fig:explainer} (Right). Given current state $X_t$, a high-level goal $G$, and action proposals $a^1...a^k$, we first use the world model to predict $z_{t+1}^1...z_{t+1}^k$, and then use a value model to obtain corresponding scores  $v_{t+1}^1...v_{t+1}^k$. Finally, we use $\arg\max$ to select the action $a_{t}$. In our setup, the action proposal model and value model are implemented using VLMs. For the value model, we first query text descriptions $y\sim p(y|z)$ and pass $y$ alongside a high-level goal $G$ to obtain value scores.

\subsection{\bench}
\label{sec:benchmark}

We can evaluate a semantic world model by measuring how well the latent $z_{t+1}$ captures the high-level semantics of the state transition $X_t\rightarrow X_{t+1}$. This can be assessed in two ways: (1) how well the text $y \sim p(y \mid z_{t+1})$ describes the observed state transition in pixel space, and (2) how accurately the model estimates the likelihood $p(o^i_{t+1}\mid q^i,z_{t+1})$ that statements $q^i$ hold in the next state. These two aspects are measured using two separate tasks: Next-State-Generation and Next-State-QA. 

\subsubsection{Task Definitions.}

\begin{figure*}
    \centering
    \includegraphics[width=1.0\linewidth]{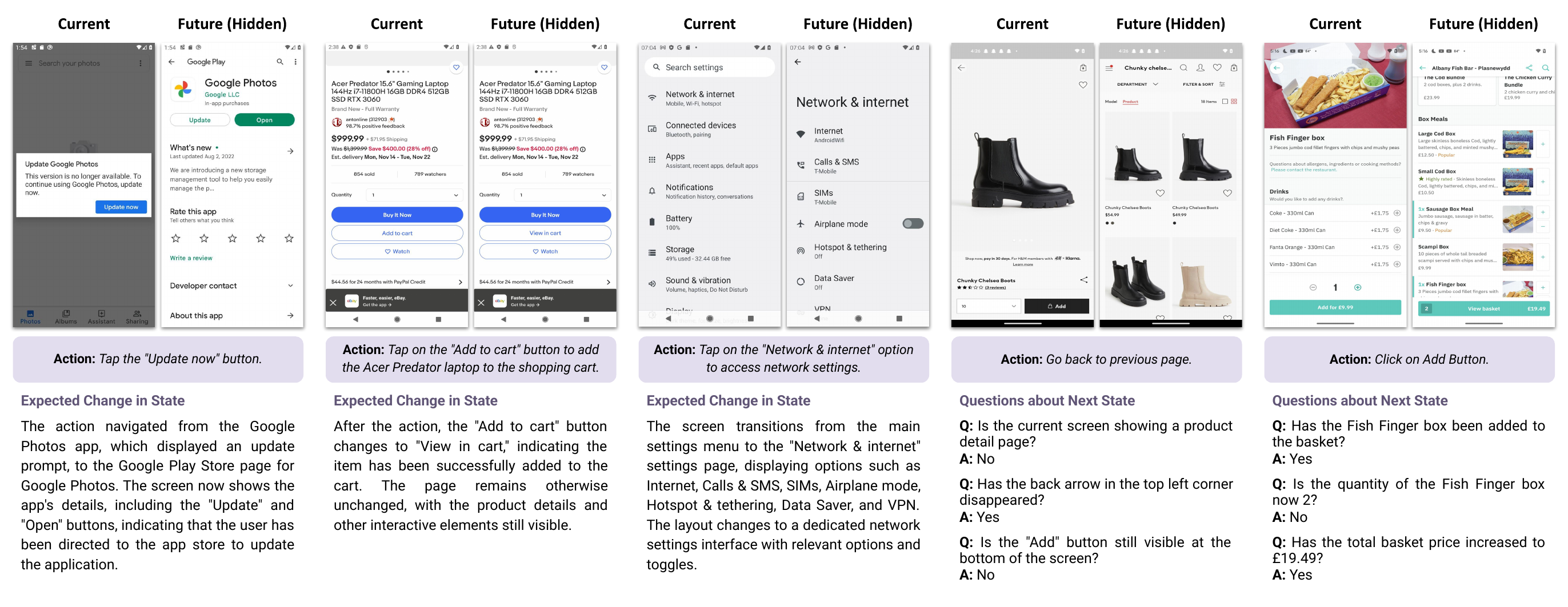}
    \caption{\textbf{Examples from \ours Training Set.} We show qualitative examples of Next-State-Generation and Next-State-QA tasks in the training set of \ours, highlighting the effectiveness of our data pipeline. }
    \label{fig:demo}
\end{figure*}

\label{subsubsec:benchmark_tasks}
\textbf{Next-State-Generation} tasks prompt a semantic world model to generate text descriptions $y_{t+1}$ describing the expected state changes given inputs $(X_t,a_t)$. We then use a VLM judge model (GPT-4o) to evaluate these predictions. The judge model is provided with inputs $X_t,a_t,y_{t+1}$, and ground-truth next state $X_{t+1}$. It then rates the quality of $y_{t+1}$ based on three key metrics:
\textbf{Accuracy.} Does the prediction accurately describe the changes in GUI states from $X_t$ to $X_{t+1}$? A model is penalized on this metric if it outputs objectively false statements.
    \textbf{Relevance.} Does the prediction involve changes that are relevant to the action being performed? A model is penalized for correct but irrelevant predictions such as ``the system time will advance." 
    \textbf{Completeness.} Does the prediction involve a detailed description of all relevant changes? 
    

For each metric, the judge assigns a score in the range of 0-5 for each output $y_{t+1}$. The overall score for Next-State-Generation is the sum of all three metrics in the range of 0-15.

\textbf{Next-State-QA} is a visual question answering (VQA) task that poses yes-or-no questions about the future state $X_{t+1}$ given current state $X_t$ and action $a_t$. This setup differs from existing GUI understanding and grounding tasks, which focus only on the \textit{observed} GUI state or user interactions. We report answer accuracy as the evaluation metric for this task.


\subsubsection{Data Generation and Filtering Pipeline}

Our pipeline for creating \bench~involves three steps: trajectory sourcing, VLM annotation, and quality-based filtering. 

\textbf{Trajectory Sourcing.} To build \bench, we source trajectories from the test split of the AndroidControl dataset~\cite{li2024effects}, which contains human demonstration trajectories with $(X_t,a_t,X_{t+1})$ triplets. The actions $a_t$ come in the form of natural language descriptions, \eg \textit{``Click on the OK option"}. We also source trajectories from Android in the Wild (AiTW) \cite{rawles2023androidinthewild}, which contains low-level user actions such as \textit{``Click (233, 324)"}. In total, 250 $(X_t,a_t,X_{t+1})$ triplets are sampled for the Next-State-Generation and 500 triplets are sampled for Next-State-QA.

\textbf{VLM Annotation.} Since some trajectories only contain low-level actions, we employ a frontier VLM (\qwenmax~) to convert these low-level actions to high-level action descriptions. We find that naively passing in pixel coordinates leads to low performance, as the model struggles to correctly interpret these coordinate values. Instead, we create visualizations of actions by overlaying markers indicating the action on the screen \cite{yang2023setofmark}. We provide additional processing details in the Appendix. 

After we obtain high-level actions for all samples, we proceed to create QA proposal. We provide $X_t,a_t, X_{t+1}$ to GPT-4o and prompt the model to propose 8 question-answer pairs based on the observed differences between $X_t$ and $X_{t+1}$, this leads to 4,000 question-answer candidates.
\begin{figure*}[ht]
    \centering
    \includegraphics[width=0.93\linewidth]{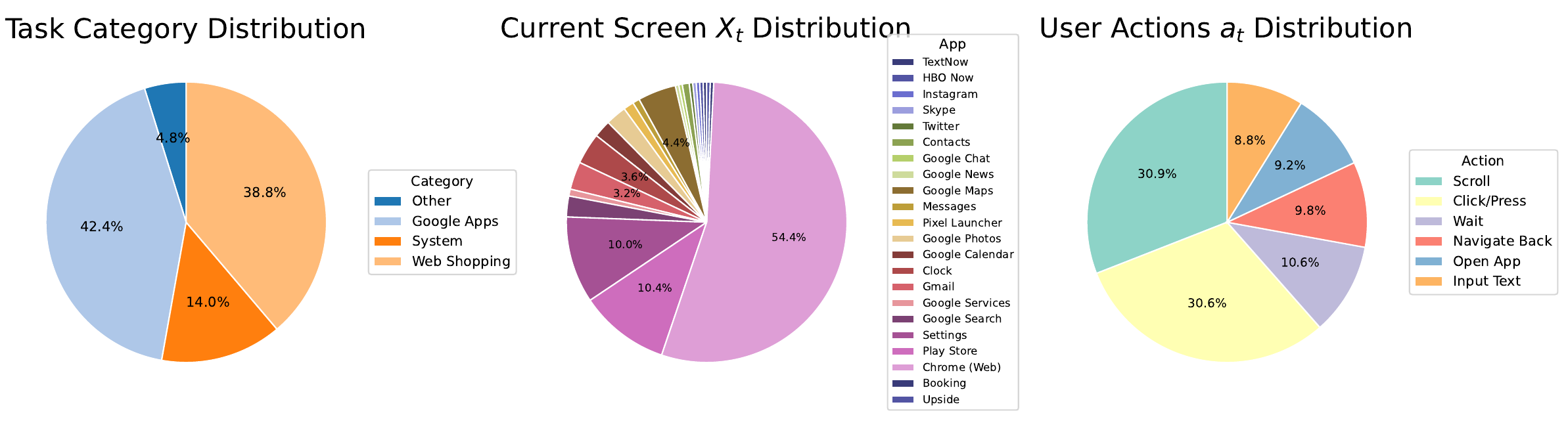}
    \caption{\textbf{Distribution of Tasks, Apps, and Actions in \bench.} \bench covers a wide range of task categories, Apps, and action types. We visualize their distributions for reference.}
    \label{fig:task_dist}
\end{figure*}

\textbf{Quality-based Filtering.} We first conduct human verification of the quality of high-level action descriptions. We find that these descriptions are of high quality and there are no obvious errors in all generated examples.  After this step, the 250 samples for Next-State-Generation task can be used without further filtering, since the trajectories are sampled from real human demonstrations and the quality of action annotations has been checked. For Next-State-QA  tasks, we heavily filter the machine-generated QA pairs to ensure only meaningful, non-trivial questions are retained. The filtering process consists of three steps. First, we prompt GPT-4o to answer its own questions by providing it with $(X_t,a_t, X_{t+1})$ triplets, and removing the questions that it fails to answer even with ground truth images (Self-Check).  Second, we use GPT-4o as a judge to filter out questions about irrelevant GUI elements such as system time, battery level, \etc (unless the element is directly relevant to the user action). Finally, we perform human verification on the remaining question-answer pairs. Human evaluators are instructed to provide a ground truth answer for the question based on  $(X_t,a_t, X_{t+1})$, and asked to determine whether the question is relevant. We remove all instances where human answers disagree with the GPT-4o annotation and remove examples that are deemed irrelevant by annotators. The final filtered QA dataset contains 1,787 questions.

\subsubsection{Coverage of Tasks, Apps, and Actions. } 

We followed the task distribution of AiTW and Android Control to ensure fair coverage. This is visualized in \cref{fig:task_dist}. Specifically, each $(X_t,a_t,X_{t+1})$ comes from a multi-turn human demonstration whose tasks fall into four broad categories: Google Apps (e.g. Mail, News, Calendar),  System (e.g. install), Web Shopping (e.g. Amazon) and Other third-party applications. The sampled screenshots $X_t$ can also come from 22 distinct apps, including Google Apps such as Google Maps and third-party apps like HBO Now, Skype, Twitter (X). We note that there is no strict mapping between task categories and Apps, and the distinction of Apps is not a subdivision of more general task categories. We also visualize the distribution of user actions $a_t$, which include common mobile GUI interactions such as scroll, click, wait, etc. 


\subsection{\ours}
\label{sec:mobileworld_method}

Our second contribution is \ours, a large foundational dataset for training semantic world models, which consists of 1.4M samples.  It includes state transition triplets $X_t,a_t,X_{t+1}$ sourced from human demonstrations, and text descriptions $y_{t+1}$ describing the changes between $X_t$ and $X_{t+1}$ resulting from action $a_t$, as well as question-answer pairs about $X_{t+1}$.

\textbf{Trajectory Sourcing.} We source triplets of $(X_t,a_t,X_{t+1})$ from human demonstrations in AiTW and the Android Control dataset largely following the same pipeline as \bench described in \ref{sec:benchmark}. Unlike \bench, we source from the training split of these datasets. 

\textbf{Annotations.} We obtain raw text descriptions of state changes by prompting a VLM to describe the observed differences between $X_t$ and $X_{t+1}$. We also obtain question-answer pairs by prompting the LLM to generate them based on observed state changes. We generate 3 text descriptions and 8 question-answer candidates per state transition. Unlike the \bench benchmark, we do not use GPT-4o as our VLM due to cost concerns. Instead, we use a strong open-sourced model \qwenmax~and Qwen3-VL-8B to generate annotations. 90\% of the data is annotated using the 8B model, while 10\% of the data is annotated using the 235B  model. We denote these splits as ``pretraining" and ``finetuning split".

\textbf{Post-processing.} For each state transition $X_t\rightarrow X_{t+1}$, we use VLM-as-a-judge to pick the best of three candidate text descriptions using the same criteria (accuracy, completeness, relevance) as \bench. We also apply the same VLM filtering techniques for question-answer pairs. The final filtered dataset consists of 543k question-answer pairs and 942k state change descriptions. Due to cost concerns, the training set is not filtered by humans.

\textbf{Task Coverage.} The task coverage and distribution of categories is similar to that of \bench (\cref{fig:task_dist}). We provide the full distribution of our training set in the Appendix.

\textbf{Quality Assessment.} To test the quality of the training data pipeline, we run the two pipelines on the test set of Next-State-Generation task. Unlike standard evaluation, our data pipeline also allows the model to access ground truth next state images. We report the same benchmark scores, including accuracy, completeness, and relevance metrics.  Further, we conducted human verification on a subset, collecting 1,800 human responses. Results show an error rate of 13.8\% before our automatic filtering pipeline and 4.7\% after filtering. 

In summary, we conducted rigorous quality control procedures when building \bench and \ours. \bench~is fully human-verified, and \ours~employs automatic quality filtering that is validated by humans on a small subset.

%







\section{Experiments}

We conduct extensive experiments to evaluate our benchmark, training dataset, and the proposed semantic world modeling paradigm.

\textbf{Evaluation.} We evaluate state-of-the-art VLMs on the Next-State-Generation and Next-State-QA tasks, including closed-source models \cite{comanici2025gemini,anthropic2025introducing_claude45,anthropic2024introducing_claude40,openai2024gpt4o} and open-source models  \cite{zhu2025internvl3,yang2025qwen3}.


\textbf{Model Training.} To further evaluate if the proposed training data can be used to improve model performance, we finetune Qwen3-VL-8B-Instruct using the \ours dataset. Further details can be found in the Appendix.


\newcommand{\pos}[1]{\textcolor{green!60!black}{#1}}
\newcommand{\negval}[1]{\textcolor{red!70!black}{#1}}

\begin{table*}[h!]
\centering
\caption{\textbf{Quantitative Results on Next-State-Generation Tasks.} We report the performance of frontier VLMs and our finetuned model. We also reported the performance of our training data annotation pipeline (\textit{Annotator}.)  }
\label{tab:main_gen}
\resizebox{1.0\linewidth}{!}{
\begin{tabular}{lcccccccc}
\toprule 
& \multicolumn{4}{c}{\textbf{Per Category Score}} & \multicolumn{3}{c}{\textbf{Breakdown Score}}&  \\
\cmidrule(lr){2-5}\cmidrule(lr){6-8} 
\textbf{Name} & \textbf{General} & \textbf{Google Apps} & \textbf{System} & \textbf{Web Shopping} & \textbf{Accuracy} & \textbf{Completeness} & \textbf{Relevance} & \textbf{Overall}  \\
\midrule
\multicolumn{9}{c}{\textit{With GT Next State Image}} \\
           Qwen3-VL-235B-A22B-Annotator &   13.17    &    13.12  &  12.34    &     13.30   &   4.45     &     4.02     &  4.61        &  13.08 \\
        Qwen3-VL-8B-Instruct-Annotator  &  13.58       & 13.09    &12.60     &    13.09&      4.42        &  4.02     &  4.61         & 13.05 \\ 
\midrule
\multicolumn{9}{c}{\textit{Without GT Next State Image}} \\
Intern-VL3-78B \cite{zhu2025internvl3} & 12.92 & 11.96 & 11.26 & 11.70 & 4.02 & 3.41 & 4.38 & 11.81 \\
Qwen3-VL-235B-A22B \cite{yang2025qwen3}& 13.08 & 12.34 & 11.09 & 12.23 & 4.11 & 3.53 & \textbf{4.51} & 12.16 \\
Gemini-2.5-Flash \cite{comanici2025gemini} & 13.08 & \textbf{12.56} & 10.77 & 11.91 & 4.14 & 3.48 & 4.45 & 12.08 \\
Gemini-2.5-Pro \cite{comanici2025gemini} & 12.92 & 11.96 & 11.23 & 12.16 & 4.04 & 3.49 & 4.46 & 11.98 \\
Claude-Sonnet-4.5 \cite{anthropic2025introducing_claude45} & 12.75 & 12.43 & 11.03 & 12.06 & 4.14 & 3.49 & 4.48 & 12.11 \\
Claude-Sonnet-4 \cite{anthropic2024introducing_claude40} & \textbf{13.25} & 12.26 & 11.20 & 12.03 & 4.07 & 3.59 & 4.41 & 12.07 \\
GPT-4o \cite{hurst2024gpt} & 12.75 & 12.38 & 11.23 & 11.77 & 4.11 & 3.44 & 4.44 & 12.00 \\
\midrule
Qwen3-VL-8B-Instruct \cite{yang2025qwen3} & 12.83 & 12.26 & 10.80 & 11.64 & 4.04 & 3.42 & 4.38 & 11.84 \\
\rowcolor{gray!20}
+SFT (Ours) & 12.83 & 12.40 & \textbf{11.63} & \textbf{12.61} & \textbf{4.19} & \textbf{3.70} & 4.50 & \textbf{12.39} \\
\rowcolor{gray!20}
$\Delta$\% (\textit{vs Baseline}) & \textbf{\pos{+0.0\%}} & \textbf{\pos{+1.1\%}} & \textbf{\pos{+7.7\%}} & \textbf{\pos{+8.3\%}} & \textbf{\pos{+3.7\%}} & \textbf{\pos{+8.2\%}} & \textbf{\pos{+2.7\%}} & \textbf{\pos{+4.7\%}} \\
\bottomrule
\end{tabular}}
\end{table*}

\begin{table}[t!]
\centering
\caption{\textbf{Quantitative Results on Next-State-QA Tasks.} We report the performance of  VLMs and human evaluation. Acc-TestMini refers to the result on the TestMini subset with 500 randomly chosen samples. }
\label{tab:main_qa}


\scriptsize
\setlength{\tabcolsep}{12pt}
\begin{tabular}{lHcHc}
\toprule
\textbf{Name} & \textbf{Params.}  & \textbf{Acc-Full} & \textbf{Count}  & \textbf{Acc-TestMini} \\
\midrule

\multicolumn{5}{c}{\textit{With GT Next State Image}} \\
GPT-4o &- & 100.0& &100.0\\
Human& -& 100.0 &&100.0\\
Qwen3-VL-8B-Instruct& 8B  & 94.9 & & 93.2 \\
\midrule
\multicolumn{5}{c}{\textit{Without GT Next State Image}} \\
Human & - & 84.4 & - & 89.2
\\

Intern-VL3-78B          & 78B  & 61.0 & 1787 &   61.8  \\
Qwen3-VL-235B-A22B          & 235B & 65.1 & 1785  & 66.2 \\
Gemini-2.5-Flash       & -    & 76.9 & 1787 & 77.6 \\
Gemini-2.5-Pro         & -    & 79.1 & 1787 & 79.0 \\
Claude-Sonnet-4.5 
                                                & -    & 71.7 & 1787 & 72.8\\
Claude-Sonnet-4 
                                                & -    & 70.3 & 1787 & 72.8 \\
GPT-4o                    & -    & 66.0 & 1787 & 69.6 \\
\midrule
Qwen3-VL-8B-Instruct       & 8B   & 67.3 & 1787 & 67.6\\
\rowcolor{gray!20}
+SFT (Ours)                                    & 8B   & 71.4 & 1787 & 74.8  \\
\rowcolor{gray!20}
$\Delta$\% (\textit{vs Baseline})               & -    & \textbf{\pos{+4.1}} & & \textbf{\pos{+6.4}} \\
\bottomrule
\end{tabular}
\end{table}

\subsection{Main Results on \bench}

We report evaluation results for Next-State-Generation (\cref{tab:main_gen}) and Next-State-QA (\cref{tab:main_qa}) tasks, and highlight key insights below.

\textbf{Existing models have considerable room for improvement.} We observe a significant performance gap between the best and worst models on both benchmarks. Notably, we find the best-performing model on the Next-State-QA benchmark, Gemini-2.5-Pro, is also one of the worst performing models on the Next-State-Generation benchmark. After careful inspection, we find that Gemini-2.5-Pro tends to generate long outputs with highly detailed, hallucinated descriptions of future states, leading to low accuracy scores. However, Gemini-2.5-Pro also has one of the highest relevance (4.48) and completeness (3.49) scores. This finding suggests that there are tradeoffs between different metrics, and Gemini-2.5-Pro prioritizes completeness and relevance over accuracy. We hypothesize that this behavior might emerge from its reasoning finetuning, as Gemini-2.5-Flash from the same family achieves a much higher generation score (+0.10) while maintaining strong QA accuracy.
 
\textbf{Training Data Pipelines Yield High-Quality Annotations.} In \cref{tab:main_gen}, both the pretraining data pipeline, which uses Qwen3-VL-8B-Instruct as the annotator, and the finetuning pipeline, which uses \qwenmax~as the annotator, outperforms all existing models, suggesting that the pipeline was able to utilize the provided next-state observations $X_{t+1}$ to generate high-quality descriptions of state transitions. These improvements in terms of accuracy and completeness scores are more pronounced than the improvements in relevance score, which is to be expected. Most notably, the gap between the pretraining and finetuning data pipeline is relatively small, suggesting that Qwen3-VL-8B-Instruct has strong enough GUI understanding capabilities to serve as a high-quality annotator.

\textbf{Finetuning on \ours improves performance on both tasks.} We find that our finetuned Qwen3-VL-8B-Instruct improves considerably on both generation and QA tasks, highlighting the utility of the \ours dataset. Notably, on generation tasks, the finetuned model achieves the highest accuracy (71.4) score and overall score (12.39).

\begin{table}[t]
\centering
\caption{\textbf{Online Evaluation on AndroidWorld.}}
\label{tab:android_world_sr}
\begin{tabular}{l c}
\toprule
\textbf{Model} & \textbf{SR} \\
\midrule
Qwen3-VL-235B-A22B & 46.9 \\
+Value Function for Actions & 49.1 \\
+Semantic WM (Zero Shot) & 50.8 \\
\rowcolor{gray!20}
+Semantic WM (Ours) & 54.3 \\
\hline
Gemini-2.5-Pro & 63.4 \\
\rowcolor{gray!20}
+Semantic WM (Ours) & 66.1 \\
\bottomrule
\end{tabular}
\end{table}

\begin{table}[t]
\centering
\caption{\textbf{Zero-shot Generalization on WebVoyager.}}
\label{tab:web}
\begin{tabular}{l c}
\toprule
\textbf{Model} & \textbf{SR} \\
\midrule
WebVoyager (GPT-4o) & 45.4 \\
\rowcolor{gray!20}
+Semantic WM (Ours) & 49.3 \\
\bottomrule
\end{tabular}
\end{table}
\subsection{Online Evaluation.}

Recent mobile GUI agents often rely on intricate scaffolds (\eg memory, RAG) to maximize performance. To isolate the contribution of semantic world models, we conduct a simple experiment comparing agents with and without a semantic world model. We use M3A \cite{rawles2024androidworld} as the base agent with \qwenmax~and Gemini-3.5-Pro the VLM. We evaluate four setups: (1) no world model, (2) directly prompt the VLM to score the actions without explict next-state prediction (3) using the pretrained VLM as both policy and world model, and (4) our finetuned Qwen3-VL-8B-Instruct as the world model. For (3)(4), we implement a model-based policy agent using the semantic world modeling framework discussed in \cref{sec:formulation}. We report task success rates in Table \ref{tab:android_world_sr}. We find that using either world model outperforms the M3A baseline, with our fine-tuned world model leading to the best performance. 

We also generalize this setup to web-navigation in a zero-shot setting and report results on WebVoyager \cite{he2024webvoyager} in Table \ref{tab:web}. At the time of experiments, 4 websites triggers captcha verification and cause the agent to fail. For fair comparison, we reproduced the baseline and report results in Table \ref{tab:web}. Additional details of these experiments are included in the appendix.

\begin{figure}
    \centering
    \includegraphics[width=1.0\linewidth]{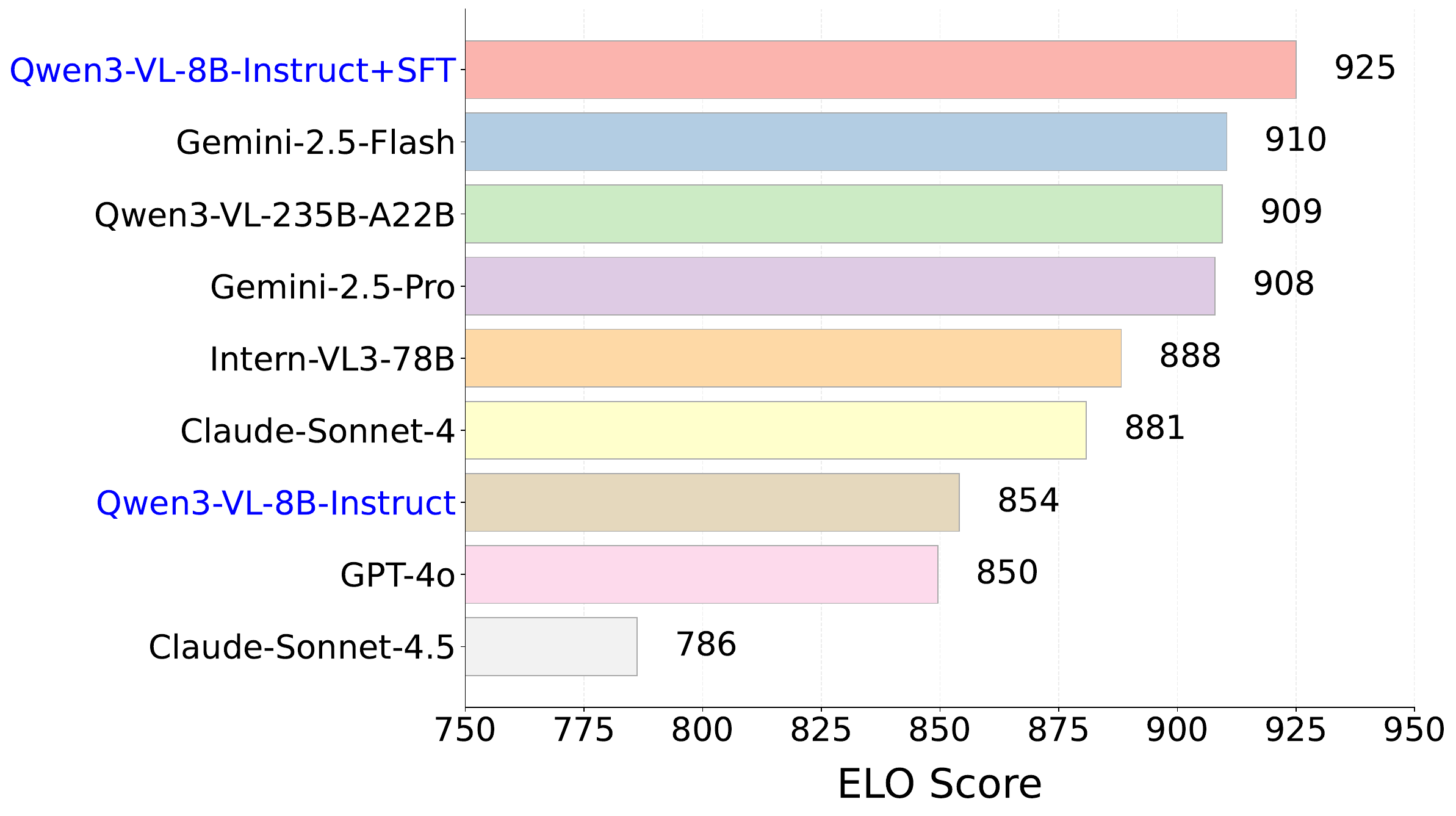}
    \caption{\textbf{Model ELO ratings from human evaluation.} }
    \label{fig:elo_scores}
\end{figure}



\subsection{Human Evaluation.}

To further verify the performance of our finetuned model, we conduct an LM-Arena-style \cite{chiang2024chatbotarenaopenplatform} human evaluation over 3,000 randomly sampled ``matches" between models. Each match pairs outputs from two different models for the same question from the Next-State-Generation task of \bench. We ask a human judge to pick a winner based on how helpful the model outputs are for decision-making. The human judge also has access to ground truth next state images $X_{t+1}$. We report ELO scores which assigns a numerical score to each model based on match results, and is widely used to evaluate LLMs, VLMs, image generators, and other types of generative AI models \cite{chiang2024chatbotarenaopenplatform,jiang2024genai,chi2025copilot}. These results are shown in \cref{fig:elo_scores}. Overall, human evaluation shows a considerable performance gain after finetuning on \ours. Further, we also verified the correlation between GPT-4o judge scores in Table \ref{tab:main_gen} and human preferences. We find that the accuracy score achieves 83\% acc in predicting human preferences. Human preference is less correlated to completeness scores

\section{Conclusion.}
In this work, we advocate for adopting semantic world models that predict state transitions at a higher level of abstraction than pixel-based world models. To support this paradigm, we introduce \bench, a high-quality, human-verified benchmark that evaluates semantic world models’ ability to accurately predict future states. We also present \ours, a large-scale dataset for training semantic world models, and demonstrate that training on this dataset leads to substantial performance gains. Finally, we showcase that semantic world models can be effectively integrated into mobile agents through a simple model-based policy, seamlessly translating world-modeling capabilities into task success in real-world environments. 

\section{Limitation}

Our work has limitations. For instance, both \ours and \bench consist solely of human demonstrations on Android, as there iOS is a close-sourced systems with complicated legal implications. We note that this limitation is common in mobile agent literature.

\section{Ethic Considerations}
Use of human annotators are approved and reviewed following respective procedures of institution and local law jurisdictions. All human annotators are properly compensated according regional income level.


\section{Acknowledgement}

AG would like to acknowledge the support from Schmidt Sciences and NSF Career Award \#2341040. SL is in part supported by Amazon Fellowship.


\bibliography{main}



\appendix
\clearpage
{

        \centering
        \Large
        \vspace{0.5em}Supplementary Material \\
        \vspace{1.0em}

}


\section{Additional Technical Details}

\subsection{Dataset Statistics}

\textbf{Comparison with Other Datasets}. We compare the size and task coverage in \cref{tab:mobile_datasets}. As shown, \ours~is the first dataset focusing on next-state prediction tasks. Its scale is also larger than most of existing datasets. We emphasize that while some datasets such as GUI-World also involves questions about multi-step trajectories, they still fall in the category of GUI understanding tasks and are fundamentally different from our world-modeling next-state prediction tasks. 

As a concrete examples, GUI-World contains questions such as ``After moving the Steam window to the center, what did
the user do next in the Edge browser?" A model is expected to answer this question based on a list of video frames from screen recording, including all past and future actions. This is a different setup than world modeling, where future states are not provided to the model.

\begin{table*}[ht]
\centering
\begin{tabular}{lccHHHHHcc}
\toprule
\textbf{Dataset} & \textbf{Size} & \textbf{Sem.} & \textbf{VL} & \textbf{Video} & \textbf{Seq.} & \textbf{Cro.} & \textbf{Dyn.} & \textbf{Task Coverage} \\
\midrule
Rico \cite{deka2017rico}& 72,219 & Low & \cmark & \xmark & \xmark & \cmark & \cmark & UI Code/Layout Generation \\
Screen2Words \cite{wang2021screen2words}& 22,417 & High & \cmark & \xmark & \cmark & \xmark & \xmark & UI Summarization \\
MetaGUI \cite{sun2022meta}& 1,125 & Low & \cmark & \xmark & \cmark & \xmark & \xmark & Mobile Navigation \\
UGIF \cite{venkatesh2022ugif}& 523 & High & \cmark & \xmark & \cmark & \xmark & \xmark & GUI Parsing \& Understanding \\
AITW \cite{rawles2023androidinthewild}& 715,142 & High & \cmark & \xmark & \cmark & \xmark & \cmark & Action Selection \\
Ferret-UI \cite{you2024ferret}& 123,702 & Low & \cmark & \xmark & \cmark & \xmark & \xmark & GUI Grounding \& Understanding \\
Spotlight \cite{li2022spotlight}& 2.5M & Low & \cmark & \xmark & \cmark & \xmark & \xmark & GUI Understanding \\
GUI-WORLD \cite{chen2024gui} & 12,379 & Both & \cmark & \cmark & \cmark & \cmark & \cmark & GUI Understanding  \\
\ours (Ours) & 1.4M & Both & \cmark & \cmark & \cmark & \cmark & \cmark & Next-State Prediction \\
\bottomrule
\end{tabular}%
\caption{\textbf{Overview of existing mobile GUI datasets.} Sem. Semantic level on instructions.}
\label{tab:mobile_datasets}
\end{table*}

\begin{figure*}[h]
    \centering
    \includegraphics[width=0.93\linewidth]{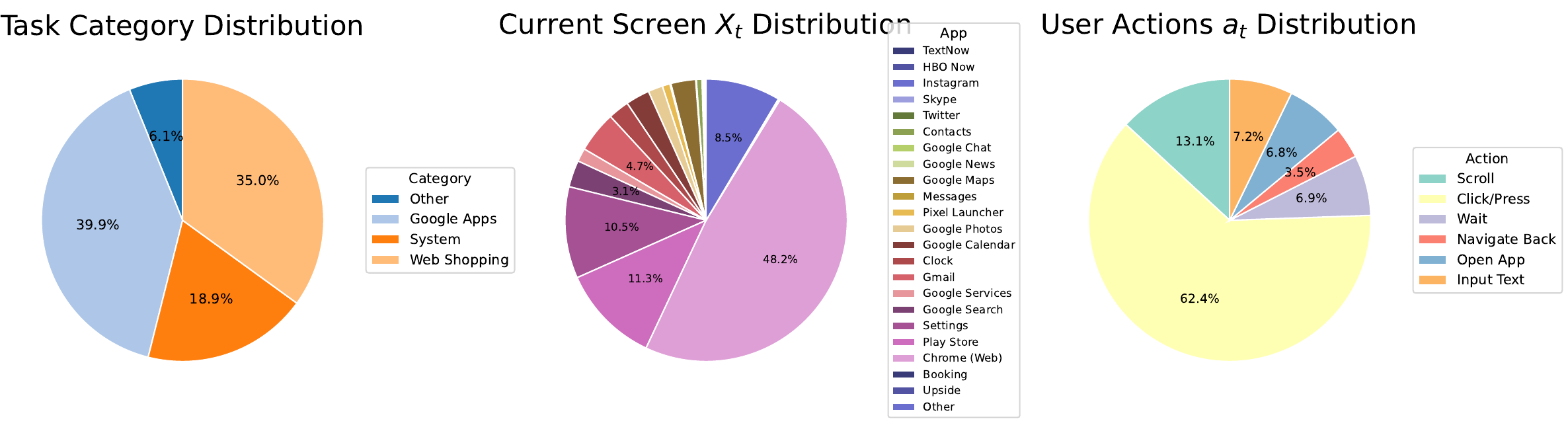}
    \caption{\textbf{Distribution of Tasks, Apps, and Actions in \ours.} \ours covers a wide range of task categories, Apps, and action types. We visualize their distributions for reference.}
    \label{fig:task_dist_train}
\end{figure*}

\textbf{Distribution of \ours}. We provided the distribution of apps and actions of \bench~in Figure \ref{fig:task_dist} of the main paper. The distribution of apps and actions of \ours~is similar to that of \bench~since they are sampled from the same source of human demonstrations. For completeness, we also include a detailed account of these distributions in \cref{fig:task_dist_train}. A major difference is that there are more click operations in the training data, which can be attributed to the differences in the train and test split of source data.

\subsection{Data Pipeline}

\subsubsection{VLM Annotation}

In this section, we provide details of the VLM annotation process. Recall that the VLM annotators serve three goals. First, they convert low level actions such as ``Click (200,312)" to high-level ones such as ``Click the return button". Second, they generate text descriptions of the visual differences in GUI state before and after an action is performed. Finally, they generate candidate QA pairs.

\begin{figure}
    \centering
    \includegraphics[width=1.0\linewidth]{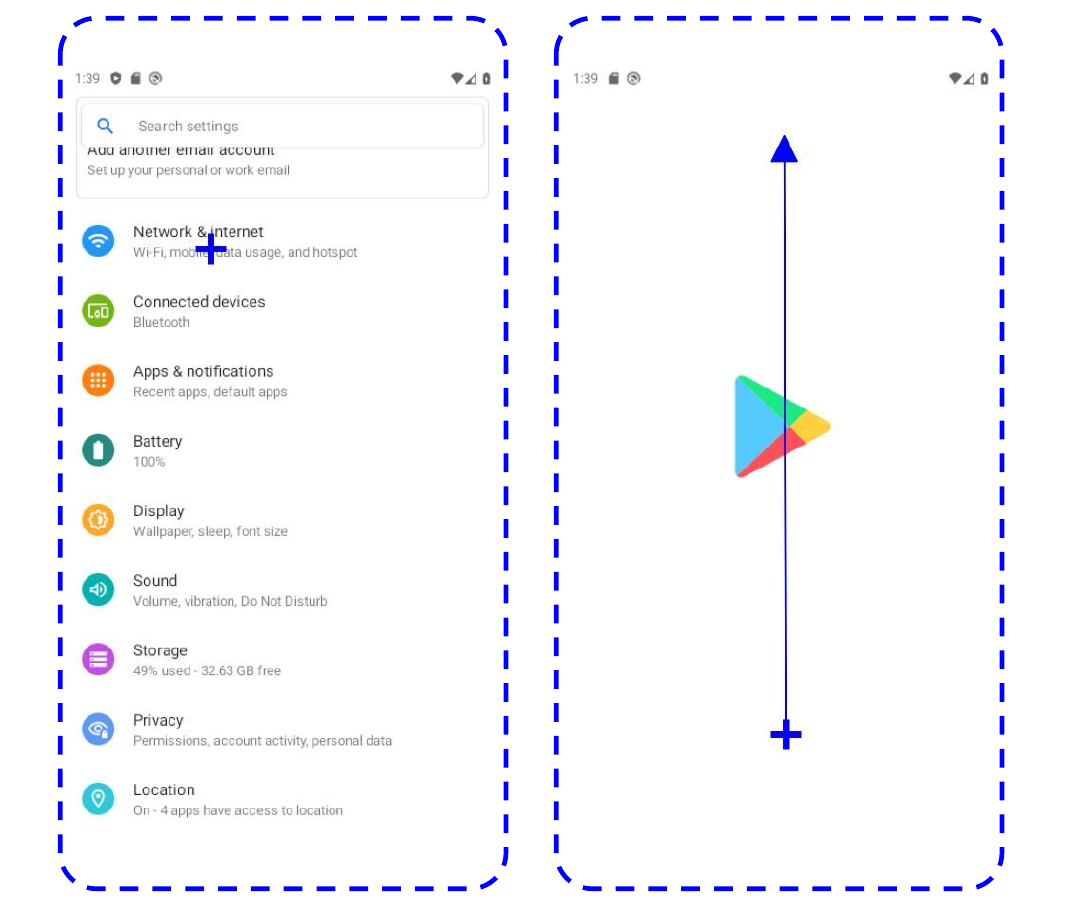}
    \caption{\textbf{Action Annotation via Visual Overlay.} To better allow VLM annotators to understand the action being performed, in addition to integer coordinates, we provide additional visual annotations on the coordinates of user gestures visually.}
    \label{fig:action_anno}
\end{figure}

\textbf{Action Representation}. Instead of only providing low level actions as text, we also annotate the image visually because we find VLMs struggle to map screen coordinates to GUI elements. An example is shown in \cref{fig:action_anno}, where we annotate click actions with a cross and swipe or scroll actions with an arrow.

\textbf{High-level Action and Change Description}. We provide a total of three images in this annotation process. Image 1 and Image 2 are screenshots of mobile phones before  a certain action. Image 2 is Image One plus the action visualization. Image 3 is the screenshot after the action is performed. The full prompt is documented as below

    \begin{tcolorbox}[breakable]
        \textbf{Annotation Prompt for High-level Action and Change Description}
        
        You are given three images which are screenshots of mobile phones before and after a certain action. Image 1 and Image 2 are states before the action. Image 2 is image one with an action annotated.

The action coordinate is marked with a blue cross. If the action is swipe, the swipe direction is marked with a blue arrow. The end of the arrow indicates the end point of the swipe.

Image 3 is the screenshot after the action is performed.

The input images have the dimension of [[Height]]x[[Width]] (Height x Width).
The action will be provided in the prompt. The overall goal will also be provided.

There are two tasks:
\begin{itemize}
    \item You should also provide a natural language description of the action based on the images.
    For example, if the raw action  provided is Tap at (200, 1112), and the image shows that the user tapped on the ``Settings" icon, you should describe the action as ``Tap on the 'Settings' icon to open the setting menu".
    If action is already descriptive enough, (e.g, Input text ``usb-c to usb-a") without any coordinate, you can just repeat it.
    Feel free to add additional context based on the image to make the action more understandable, such as Input the text ``usb-c to usb-a" in the search box to search for usb-c to usb-a adapters.
\item Your task is to describe the difference between the two images in detail, focusing on the changes that occurred as a result of the action.
    Please provide a comprehensive description of the differences, including any visual changes, layout modifications, color alterations, text changes, or any other noticeable differences.
    The overall goal is for reference only, do not include it in your answer or rely on it too much. It may not be fully accurate.
    Limit your response to one small paragraph less than 200 worlds. Only desribe the changes, no need to say which things are not changed. Also feel free to ignore system status such as wifi,battery, etc.
    If action is not successful or nothing is changed, just say No changes observed after the action. And explain wht it fails
\end{itemize}

The action is: ``\{low level action \}"

The goal is ``\{goal \}"

Your respone should be in the following format:

Action Description: [Your description of the action based on the images.]

Change Description: [Your description of the changes between Image 1 and Image 3.]
    \end{tcolorbox}

\textbf{QA proposals}. We sample 8 question-answer (QA) proposals for each $(X_t,a_t,X_{t+1})$ triplet. The prompts are documented as follows

    \begin{tcolorbox}[breakable]
        \textbf{Annotation Prompt for Creating QA Candidates}    
        You are given two images representing the before and after states of an Android device screen. The action being performed is \{action\}.

Now generate 5 QA pairs that test a model's understanding of the changes between these two images. Specifically, you should provide yes/no questions about UI states after the change, as well as ground truth answers based on the changes observed.

Do you include changes that are not directly related to the action, such as system time, battery level, etc.
For each QA pair, format it as:

Q: [question]

A: [answer]

Provide 8 such QA pairs in total.

Provida a balanced mix of true and false answers.

Use exactly the format provided, do not add any extra text or formats (e.g. bold)
    \end{tcolorbox}

Given 500 state transitions, the total number of generated QA candidates are 4,000.

\subsubsection{VLM Filtering} 

We perform additional filtering to 1) remove questions whose answer does not accurately reflect state-changes 2) remove irrelevant question-answer pairs relating to system time, battery level, etc. In particular, we find 2) is necessary even though we have instructed the model not to produce this kind of questions during the generation process, as we find that the model will occasionally ignore these instructions and create trivial or irrelevant questions regardless. 

To achieve 1) removing questions whose answer does not accurately reflect state-changes, we perform self-check verification where the model is provided with the current state, user action, and ground-truth next state to answer a question. The prompt is documented as follows:

    \begin{tcolorbox}[breakable]
        \textbf{Prompt for VLM Self-Check}
        
        You are an intelligent GUI agent capable of understanding GUIs and actions on mobile devices. Given the current GUI screenshot and input action, answer the following questions based on your predictions of the changes that will occur on the next screen after the action is performed.

The action is \{action\}

Answer with yes or no. 

The question is \{question\}. 

You will also be given ground truth next state image as image 2.
    \end{tcolorbox}

To achieve 2) removing irrelevant question-answer pairs, we employ the following prompt.

    \begin{tcolorbox}[breakable]
        \textbf{Prompt for Relevance Filter }
        
        You are given two images representing the before and after states of an Android device screen. The action being performed is \{action\}.

A question-answer proposal is given as follows:

Q: \{question\}

A: \{Answer\}

Please decide if the question asks about expected state changes that are relevant to the action. Non-relevant actions include system time, battery level, Wi-Fi status, signal strength, etc., unless these aspects are directly related to the current action (e.g. enabling airplane mode will disable cellular signal).

Note that even if a question asks about some UI elements that did not actually change, it may still be relevant. For example, when the user is typing the city column of an address form, it is ok to ask if the actions will result in any changes in the street address column. 

Answer with a simple yes-or-no.
    \end{tcolorbox}

After the filtering process, we are left with 2,458 QA pairs out of 4,000 proposals.

\subsubsection{Human Filtering}

\begin{figure*}[h!]
    \centering
    \begin{subfigure}[b]{0.48\linewidth}
        \centering
        \includegraphics[width=\linewidth]{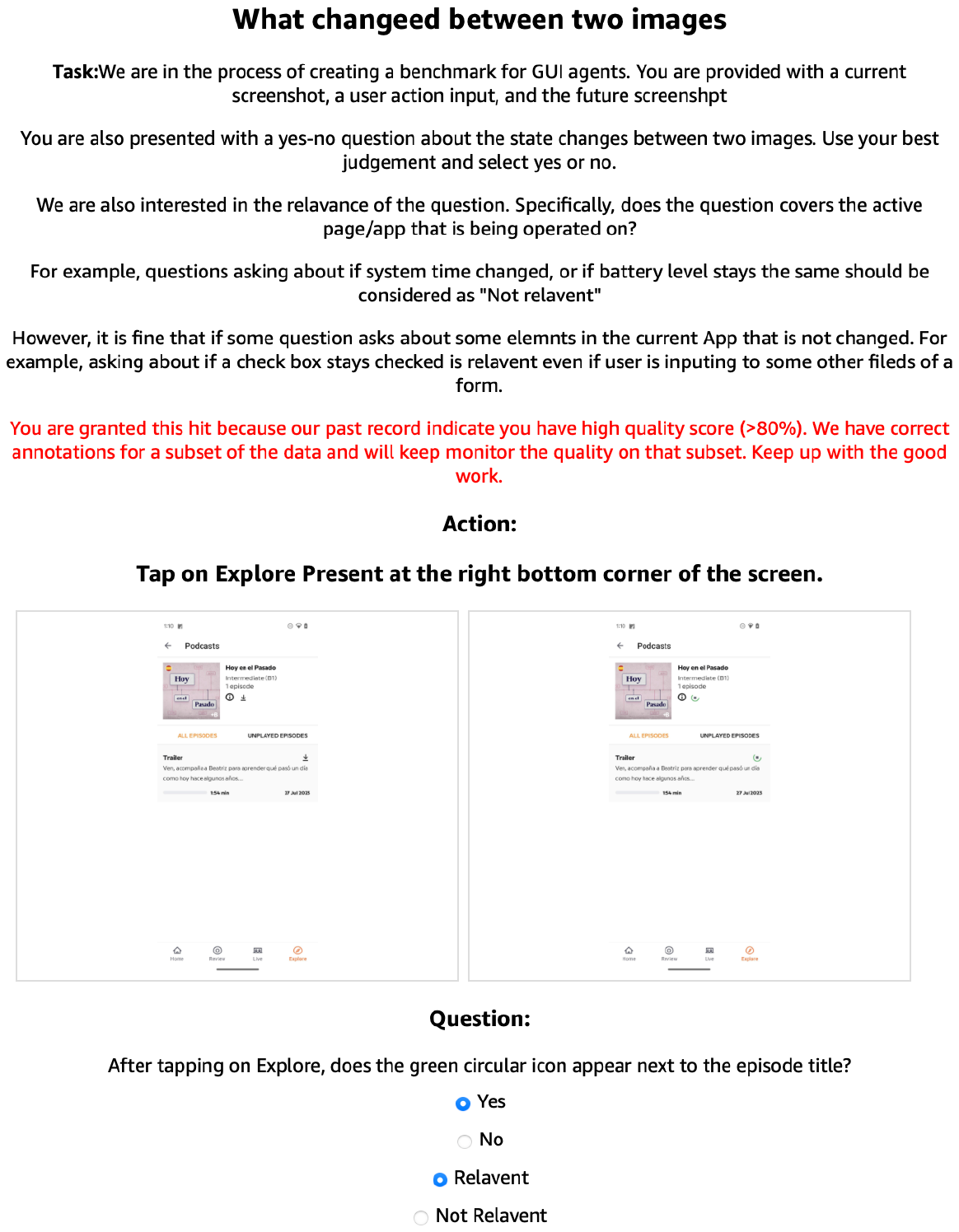}
        \caption{\textbf{Correctness and Relevance Filter.}}
        \label{fig:human_verifier_1}
    \end{subfigure}
    \hfill
    \begin{subfigure}[b]{0.48\linewidth}
        \centering
        \includegraphics[width=\linewidth]{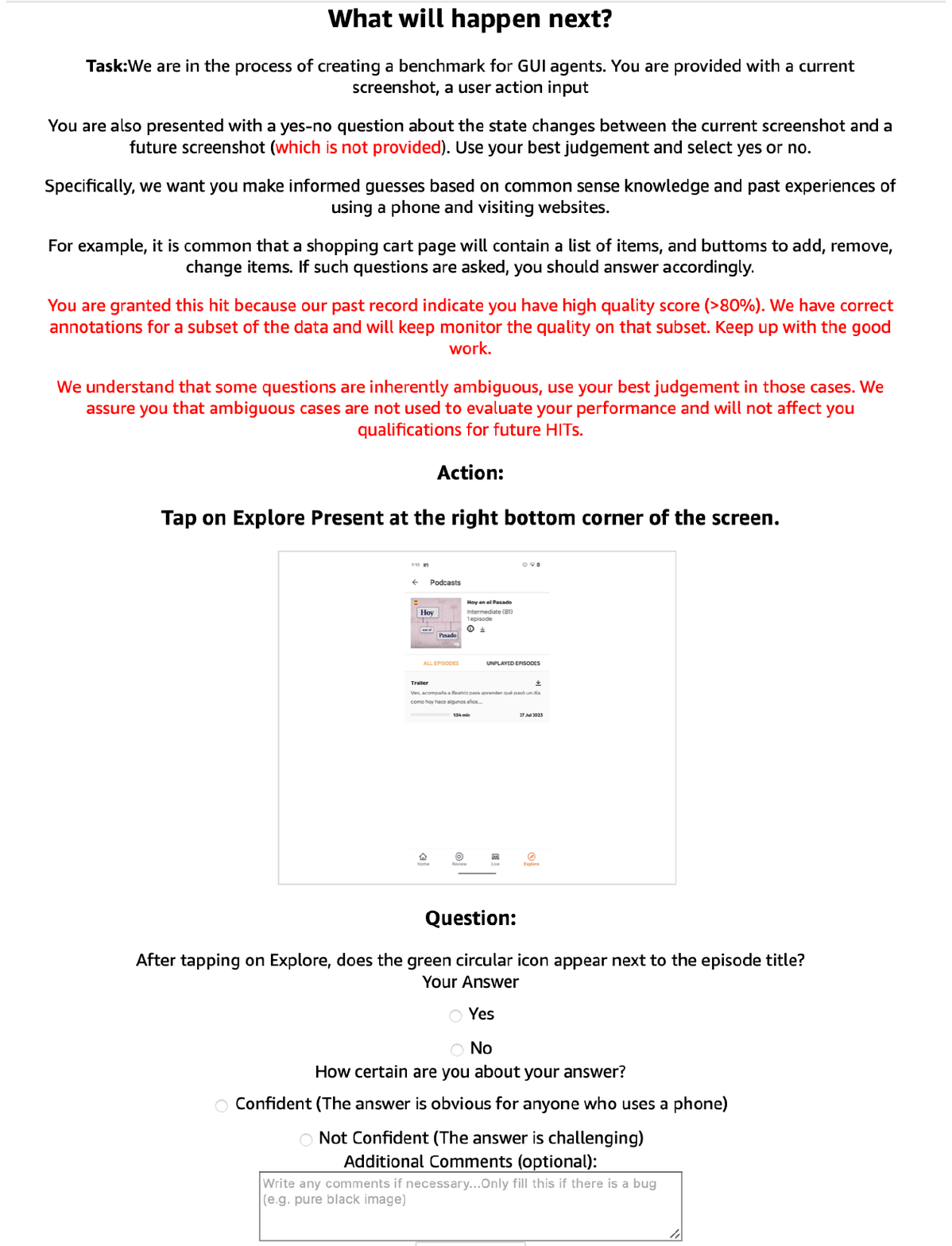}
        \caption{\textbf{Ambiguity Filter.}}
        \label{fig:human_verifier_2}
    \end{subfigure}

    \caption{\textbf{Example Screenshots of the User Interface for Human Annotators Used in the Data Filtering Process.}  
    (a) Interface for applying correctness and relevance filters.  
    (b) Interface for detecting ambiguous responses.}
    \label{fig:human_verifier_combined}
\end{figure*}

For the remaining 2,458 QA pairs, we conduct additional human filtering to further ensure the quality of the benchmark. We employ Amazon Mechanical Turk for this process. To ensure the quality of the worker, the authors annotated a small 100-sample test set manually and only employ workers with high performance on this test set to work on the full set. The human annotation process consists of the following two steps:

\textbf{Filtering Incorrect or Irrelevant QA pairs.} The purpose of this step is similar to the VLM filtering process. We ask the human annotator to judge 1) if the ground truth answer is consistent with the observed state changes observed in screenshots, 2) If the question is relevant to the action being performed. We list the prompt in Figure \ref{fig:human_verifier_1}.

\textbf{Ambiguity.} While a question can be correct and relevant, it may be ambiguous or under-defined if it asks about something that is impossible to reasonable predict. For example, when opening the ``sports" tab of a new app, it is impossible to predict what the headline will be. Hence, we additionally ask human annotators to judge if the questions can be reasonably answered by an average user.  The interface is shown in \cref{fig:human_verifier_2}

After this process, we are left with 1,787 QA pairs as our benchmark.

\section{Additional Experiment Details and Results}

\subsection{Training Setup}

We finetune the base model Qwen3-VL-8B-Instruct \cite{yang2025qwen3} on the proposed \ours~dataset for 2 epochs, with a learning rate of $2\times10^{-6}$ on 8 A6000 GPUs. The detailed training setup is listed in \cref{tab:training-stages}

\begin{table}[h]
\centering
\caption{\textbf{Training configurations on \ours.} We report the relevant hyperparameters for training, including the learning rate (LR), number of training steps, optimizer setup, image resolution for understanding and generation tasks. }
\label{tab:training-stages}
\begin{tabular}{lHHc}
\toprule

 & \textbf{Stage 1} & \textbf{Stage 2} & \textbf{SFT} \\
\midrule
Learning Rate (LLM Backbone) & $5 \times 10^{-6}$ & $1 \times 10^{-4}$ & $2 \times 10^{-6}$ \\
Learning Rate (Vision Encoder) & $5 \times 10^{-7}$ & $1 \times 10^{-4}$ & $2 \times 10^{-7}$ \\
Epochs & 80k & 400k & 2 \\
$\beta_1$ & 0.99 &0.99  & 0.99 \\
$\beta_2$ & 0.999 &0.999  & 0.999 \\
Optimizer & AdamW & AdamW & AdamW \\
\midrule

Image resolution & 1280px & 384 $\times \{(1,3),(2,2)\}$ & 1280px\\
Batch Size & - & 256 $\rightarrow$ 512 $\rightarrow$ 1024 & 128 \\
Warmup & - & 256 $\rightarrow$ 512 $\rightarrow$ 1024 & 6\% \\
LR Schedule & - & 256 $\rightarrow$ 512 $\rightarrow$ 1024 & Cosine \\
\midrule
\end{tabular}%

\end{table}

\subsection{Evaluation Setup}

\subsubsection{Model Choice and Sampling Parameters}
In this section, we document the exact model version used to reproduce our main results, as well as the sampling parameters. The model ids of closed source models are listed in \cref{tab:closed_source_map}. For open-sourced models, we use their Hugging Face releases.  For next-state-generation tasks, we use the default sampling parameters of respective models and do not pass in extra parameters. For next-state-QA tasks, we set temperature to 0.0 for reproducibility. We noticed that setting the temperature to 0.0 will lead to text repetitions in some models for next-state-generation tasks. Hence, we set temperature to 0.0 only for short QA tasks.

\begin{table*}[ht!]
\centering
\caption{\textbf{Closed-Source Model Version Mapping.} We list the precise model identifiers corresponding to each name used in Table~\ref{tab:main_gen}.}
\label{tab:closed_source_map}
\begin{tabular}{lll}
\toprule
\textbf{Name in Table} & \textbf{Specific Version / Model ID}  & \textbf{License}  \\
\midrule
Gemini-2.5-Flash & \texttt{gemini-2.5-flash (stable) }  & Google APIs Terms of Service  \\ 
Gemini-2.5-Pro & \texttt{gemini-2.5-pro  (stable)}& Google APIs Terms of Service   \\
Claude-Sonnet-4 & \texttt{claude-sonnet-4-20250514} & Commercial Terms (Claude) \\
Claude-Sonnet-4.5 & \texttt{claude-sonnet-4-5-20250929}  & Commercial Terms (Claude) \\
GPT-4o & \texttt{gpt-4o-2024-08-06 (default)} & Terms of Use
 (OpenAI) \\
\bottomrule
\end{tabular}
\end{table*}

\subsubsection{Automatic VLM Evaluation}

We employ GPT-4o as the judge model. We employ the following prompt to obtain the scores for next-state-generation results
    \begin{tcolorbox}[breakable]
        \textbf{Prompt for Automatic Evaluation }
        
        You are a judge to judge a VLMs's ability to understand the action on mobile device GUI. Given the current GUI screenshot and input action, the model will describe the changes that will occur on the next screen after the action is performed.

The action is \{action\}.

You are provided with the model's response, an input image of the current state (first image), as well as ground truth next state image (second image).

You are also given a reference text that describes the actual changes that happened after performing the action.
Your task is to evaluate the model's response based on the following criteria:

1. Accuracy: Does the model's description accurately reflect the changes that occurred in the GUI after the action?

2. Completeness: Does the model mention all significant changes that are visible in the next state image?

3. Relevance: Are all the changes mentioned by the model relevant to the action performed, or are there extraneous details?

For each criterion, assign a score from 1 to 5, where 1 is poor and 5 is excellent.

After evaluating all three criteria, provide an overall score out of 15, along with a brief justification for your scores.

The model's response is \{response\}

The reference changes are \{changes\}

Format your evaluation as follows:

----Begin of response----

Accuracy: [score]

Completeness: [score]

Relevance: [score]

Overall Score: [total score]

----End of response----

use the exact format without any additional text.
    \end{tcolorbox}

\subsubsection{Human Evaluation}
In human evaluation, we find that forcing human evaluators to give a numerical score leads to inconsistent behavior across different annotators, whose score range varies. To address this, we instead ask human evaluators to compare outputs from two models. 

\begin{figure*}[t]
    \centering
    \includegraphics[width=0.99\linewidth]{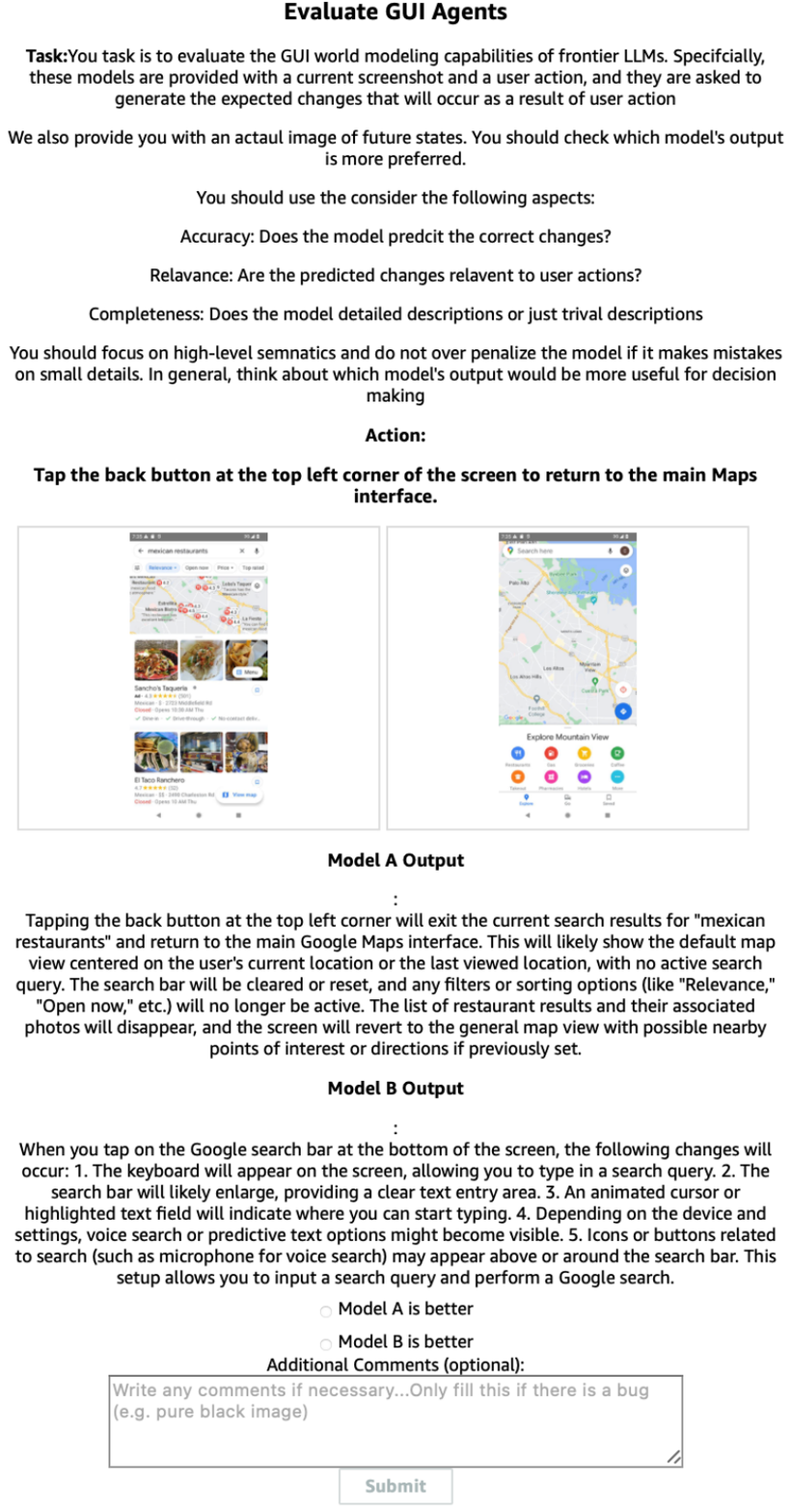}
    \caption{\textbf{Human Evaluation Interface of Comparing Two Models.}}
    \label{fig:human_elo_interface}
\end{figure*}

We illustrated the human evaluation interface in \cref{fig:human_elo_interface}. In total, we create 3,000 ``matches" by randomly selecting benchmark entries and model pairs. We compute ELO scores following the setup of LM-Arena and Gen-AI-Arena \cite{chiang2024chatbotarenaopenplatform,zhou2023webarena,jiang2024genai}

\subsubsection{AndroidWorld}

For our AndroidWorld baseline, we use M3A implemented via the official AndroidWorld codebase \cite{rawles2024androidworld}. For our model-based policy, a value model is needed to evaluate the predicted next states. We use the following prompt to obtain value scores for future states:

     \begin{tcolorbox}[breakable]
        \textbf{Prompt for Value Model}
        
        [...Omitted Background Description Copied From Action Selection Prompt of M3A ...]

Now based on the goal and expected state changes caused by the proposed 8 actions, reason about the value of the resulting states from these actions and score each of them in the range of 1-10. Higher score means the resulting state is more likely on the right trajectory to reach the final goal.

Think carefully before giving the final answer. Your Final answer should be the following format (Expected\_Change and Action should be copied from above)

Action 1: \{\{"action\_type":...\}\} Expected\_Change: ... Score\_Reason:... Score: ...

Action 2: \{\{"action\_type":...\}\} Expected\_Change: ... Score\_Reason:... Score: ...
    \end{tcolorbox}

For better presentation, we omit the long background description that describes the current environment, past action history, overall goal, and some hard-coded tips like ``Sometimes you may need to navigate the phone to gather information needed to complete the task." Note that this portion is directly copied from M3A agents.

\textbf{WebVoyger} The standard WebVoyger agent provides a VLM (GPT-4V) with current screen observation and a list of possible actions. It asks the VLM to directly select an action from the provided list and provide reasoning. We modify this process by first generate next-state predictions using our semantic world model and then run a value model using the same setup as AndroidWorld Experiment. 

In our experiemnts, we countered Captcha verification on the following websites: Google Search, Booking, Cambridge Dictionary,  Allrecipes. These four websites are removed from benchmark since encountering Captcha necessitates task failure.

\subsection{Additional Results}

\textbf{Statistic Significance Analysis.} To verify the statistic significance of our results, we report the 95\% confidence interval of next-state-generation task in Table \ref{tab:statistics}. Results show that our improvements are statistic significance (p<0.05).
\begin{table*}[h]
\centering
\begin{tabular}{lcccc}
\toprule
Model & Accuracy & Completeness & Relevance & Overall Score \\
\midrule
Qwen3-VL-8B-Instruct & $4.05 \pm 0.07$ & $3.37 \pm 0.06$ & $4.35 \pm 0.06$ & $11.77 \pm 0.19$ \\
+SFT (Ours) & $4.22 \pm 0.07$ & $3.72 \pm 0.06$ & $4.55 \pm 0.06$ & $12.49 \pm 0.18$ \\
\bottomrule
\end{tabular}
\caption{Statistics Significance analysis. To further verify the significance of our results, we conducted additional analysis by sampling 5 times and report the $95\%$ confidence interval. Due to token cost issue, we only perform this validation for open-sourced Qwen3-VL-8B-Instruct and our finetuned model. Results show that the improvements from finetuning on \ours~dataset are statisctically significant. }
\label{tab:statistics}
\end{table*}
\subsubsection{More Qualitative Examples.}

\begin{figure*}
    \centering
    \includegraphics[width=0.98\linewidth]{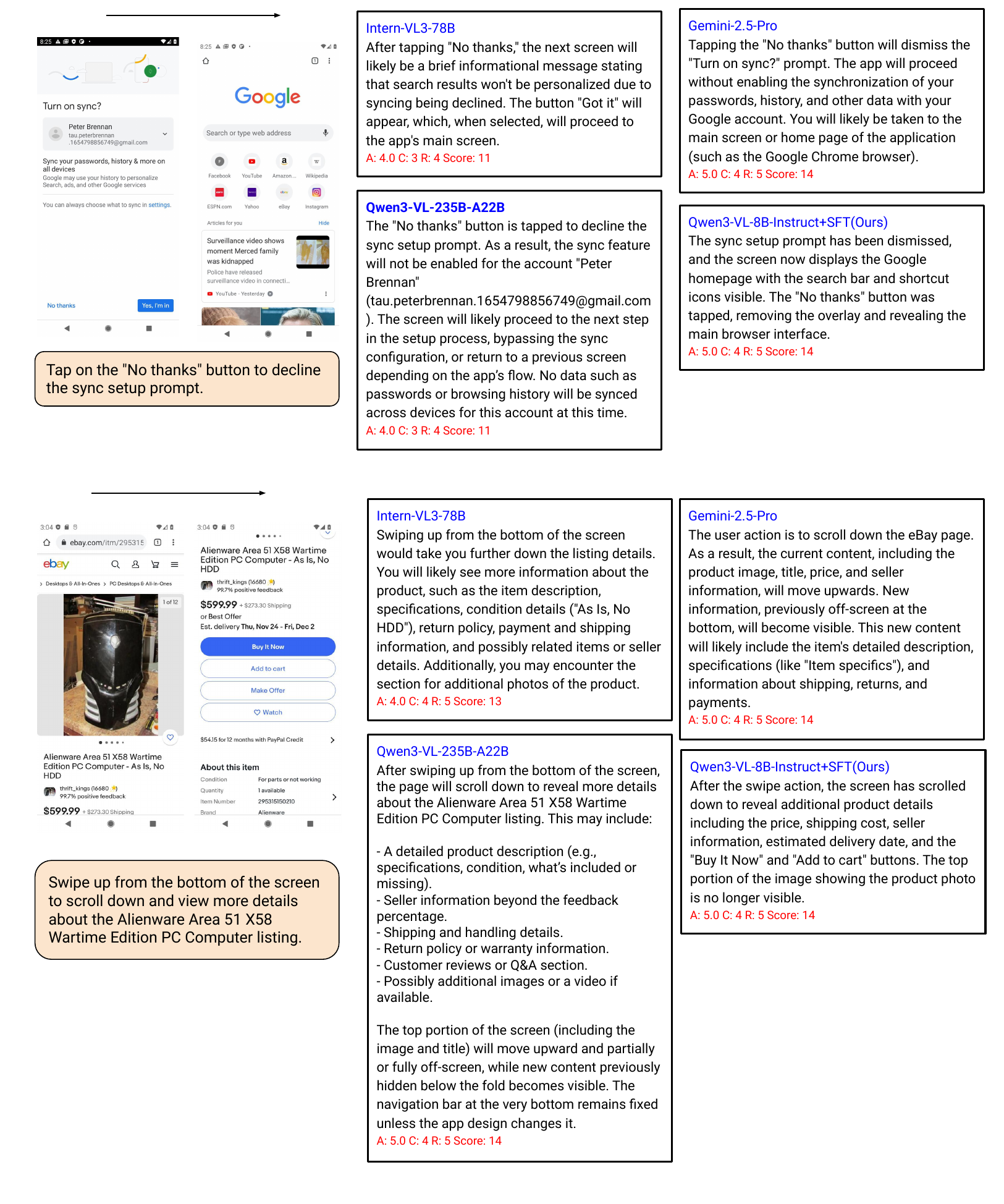}
    \caption{\textbf{Qualitative Examples of \ours~ Evaluation Results.}}
    \label{fig:qualitative-examples-appendix}
\end{figure*}

In \cref{fig:qualitative-examples-appendix}, we provide additional qualitative examples on screenshots, user actions, and model outputs on \ours, as well as the GPT judge score. Model performance varies across different samples. Our finetuned model is among the top performing models for most of the samples. 

\subsubsection{Model-Scaling}

To further analyze the difficulty of the proposed tasks and investigate the model scaling behavior, we conducted an additional experiment by finetuning a smaller model Qwen3-VL-2B-Instruct following the same setup. We report results in \cref{tab:qwen3_scaling}. We observe that while the relative improvements of finetuning is larger for the 2B model (+5.4) versus the 8B model (+4.1), the 2B model significantly underperforms the 8B model. We believe this gap is largely caused by limited GUI understanding capabilities, of the 2B model.  We verify this hypothesis by additionally evaluate the models' performance when provided with the ground truth next-state image. This process coverts next-state-QA to standard VQA tasks where the model only needs to compare the differences in the two provided images. We find that GPT4o, Human annotator, and Qwen3-VL-8B-Instruct achieves high scores in this setup ($>90$\%), while Qwen3-VL-2B-Instruct only achieves $70.0\%$ accuracy, indicating its limited capacity. These findings indicate that the proposed world modeling task is challenging and is beyond the capacity of small VLMs.

\begin{table}[h]
\centering
\caption{\textbf{Model Scaling Results.} We provide qualitative results on next-state-QA tasks of finetuned Qwen3-VL-2B-Instruct and Qwen3-VL-8B-Instruct. }
\label{tab:qwen3_scaling}
\begin{tabular}{l c}
\toprule
\textbf{Model} & \textbf{Acc} \\
\midrule
\multicolumn{2}{c}{\textit{With GT Next State Image}} \\
GPT-4o & 100.0 \\
Human& 100.0 \\
Qwen3-VL-8B-Instruct & 94.9\\
Qwen3-VL-2B-Instruct & 70.0\\
\midrule
\multicolumn{2}{c}{\textit{Without GT Next State Image}} \\
Qwen3-VL-8B-Instruct & 67.3\\
\rowcolor{gray!20}
+SFT(Ours) & 71.4 \pos{(+4.1)} \\
Qwen3-VL-2B-Instruct  & 60.6 \\
\rowcolor{gray!20}
 +SFT(Ours) & 66.0  \pos{(+5.4)} \\
\bottomrule
\end{tabular}
\end{table}


\subsubsection{Combining Semantic World Models to Pixel-Space World Models}

\begin{figure*}[h]
    \centering
    \includegraphics[width=0.93\linewidth]{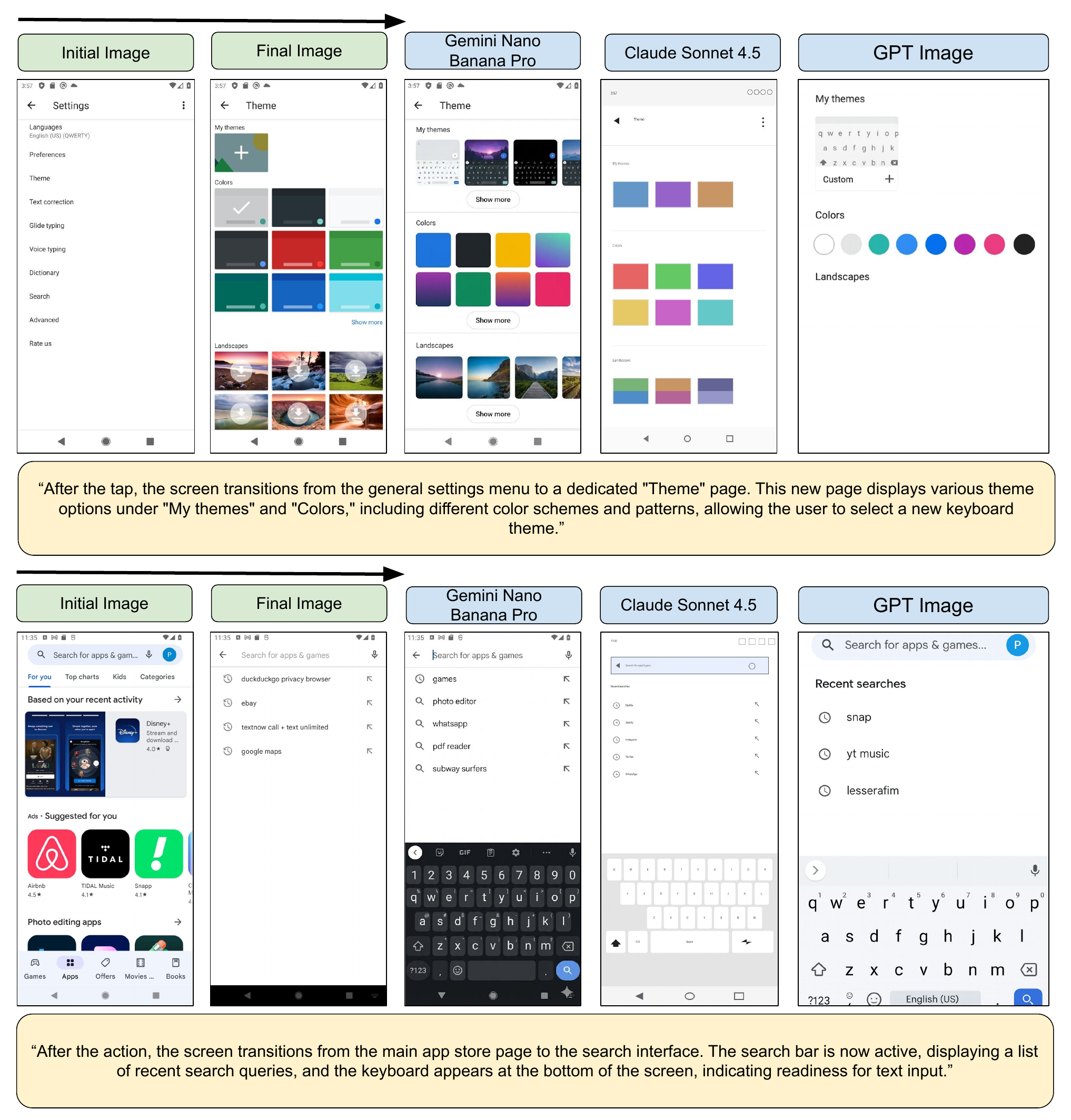}
    \caption{\textbf{Qualitative Results of Pixel Modelling} We combine the output of semantic world models with pixel renders to obtain image predictions.  Gemini Nano Banana Pro achieves the best qualitative results. We demonstrate that using text as guidance in the generation process results in a prediction much closer to the ground truth.}
    \label{fig:nano_banana}
\end{figure*}

While we argue that semantic world models are conceptually more useful and less expensive to train than pixel world models, we emphasize that conceptually semantic world models can be considered as a sub-process of the classic world models, as noted in \cref{sec:formulation}. To validate this paradigm, we can pass the output of our semantic world model to a state-of-the-art pixel renderer to achieve pixel world models. We illustrate this approach in Figure \ref{fig:nano_banana}, where we combine our finetuned VLM with frontier image generators such as Nano-Banana \cite{comanici2025gemini} to produce pixel outputs. We find that accurate text prediction can be used to create highly plausible next-state screenshots. We emphasize that this works focus on the semantic level, we left more explorations in pixel world modeling for future works.

\section{Reproducibility Statements}

All artifacts, including fine-tuned model weights, evaluation dataset, and training dataset will be open-sourced. We will also release the evaluation codebase for Next-State-Generation, Next-State-QA, and AndroidWorld experiments.

\section{Additional Discussion with Related Works.}

\textbf{ViMo.} We note that a line of literature tangent to this work,  notably ViMo \cite{luo2025vimo} focuses on building pixel-space GUI world models by addressing challenges such as text rendering and GUI consistency. We are unable to compare with ViMo because the authors did not release the model checkpoints, training data, or evaluation data.

\textbf{Language-based Abstraction for Decision-Making.} Using language-based abstractions has been widely studied for decision-making problems. WorldPrediction \cite{chen2025worldprediction} provide a theoretical framework that recognizes the world state has "task-relevant components" and "task-irrelevant" components, and suggest "change captioning" can be used as an alternative to raw image inputs.  Vision Language World Model \cite{chen2025planning} performs world modeling in language space by prediction captions of future video frames. However, most of these works focus on real physical world with embodied agents. Our work specifically focus on introducing the semantic world modeling paradigm for mobile agents. 

\section{Additional Limitations}
Our work share many common limitations of VLM-based works, such as hallucinations in the generated output. We conducted rigorous human verification for our benchmark to ensure its quality. However, such approach is not fesaible for training data due to cost issue. As discussed in Section \ref{sec:mobileworld_method}, human validation on a subset of training data showed that our automatic filtering pipeline reduced the error rate from 13.8\% to 4.7\%, but there may still be rooms for improvements for future works (e.g by using more costly models for filtering.)

\end{document}